\documentclass{article}

\usepackage{microtype}
\usepackage{graphicx}
\usepackage{svg}
\usepackage{subcaption}
\usepackage{booktabs} % for professional tables
\usepackage{array}
\usepackage{bbm}

\usepackage{hyperref}

\usepackage[preprint]{icml2026}

\usepackage{amsmath}
\usepackage{amssymb}
\usepackage{mathtools}
\usepackage{amsthm}

\usepackage[capitalize,noabbrev]{cleveref}

\theoremstyle{plain}

\theoremstyle{definition}

\theoremstyle{remark}

\usepackage[textsize=tiny]{todonotes}

\definecolor{wb}{rgb}{0, 0.5, 0}

\icmltitlerunning{Sparse Masked Attention Policies for Reliable Generalization}

\begin{document}

\twocolumn[
%\icmltitle{Sparse Attention Masks for Reliable Generalization \\ in Deep Reinforcement Learning}
\icmltitle{Sparse Masked Attention Policies for Reliable Generalization}%\\ in Deep Reinforcement Learning}



\icmlsetsymbol{equal}{*}

\begin{icmlauthorlist}
\icmlauthor{Caroline Horsch}{tud}
\icmlauthor{Laurens Engwegen}{tud}
\icmlauthor{Max Weltevrede}{tud}
\icmlauthor{Matthijs T. J. Spaan}{tud}
\icmlauthor{Wendelin B\"ohmer}{tud}
\end{icmlauthorlist}

\icmlaffiliation{tud}{Department of Intelligent Systems, Delft University of Technology, Delft, Netherlands}

\icmlcorrespondingauthor{Caroline Horsch}{cchhorsch@tudelft.nl}

\icmlkeywords{Machine Learning, ICML}

\vskip 0.3in
]



\printAffiliationsAndNotice{}  % leave blank if no need to mention equal contribution

\begin{abstract}
In reinforcement learning, abstraction methods that remove unnecessary information from the observation are commonly used to learn policies which generalize better to unseen tasks. However, these methods often overlook a crucial weakness: the function which extracts the reduced-information representation has unknown generalization ability in unseen observations. In this paper, we address this problem by presenting an information removal method which more reliably generalizes to new states. We accomplish this by using a learned masking function which operates on, and is integrated with, the attention weights within an attention-based policy network. We demonstrate that our method significantly improves policy generalization to unseen tasks in the Procgen benchmark compared to standard PPO and masking approaches. 
\end{abstract}

\section{Introduction}
\label{introduction}
Reinforcement learning methods excel at finding policies which are able to solve a wide range of complex individual tasks. However, these learned policies often fail to transfer to related tasks, even when these tasks are semantically similar or even identical up to minor visual changes. This is the Zero-Shot Policy Transfer \citep[ZSPT,][]{kirk2023survey} setting, where an agent is given a limited subset of similar tasks to train on, and is then evaluated on the performance in unseen tasks from the full set.   

One common approach to this problem is information removal in the form of feature selection, visual masking, and representation learning. All of these methods rely on the common understanding that reducing the amount of information in the state representation can reduce spurious correlations and overfitting, leading to better generalizing functions. While these methods of information removal have demonstrated vastly improved generalization \citep{bertoin2022look, grooten2023madi}, the way the removal {\em itself} is computed must generalize equally well. This presents a paradox: to reliably improve generalization of policies, the information removal must generalize reliably as well. In order to achieve reliable information removal, we propose to integrate weight masking into an attention-based architecture, which sparsifies the mask computation at the same time as the policy, removing unnecessary input information and thus improving the generalization of both.

\paragraph{Sparse-Dependent Policies}
\label{sparse-dep-policies}

\begin{figure}
    \centering
    \includegraphics[width=0.4\linewidth]{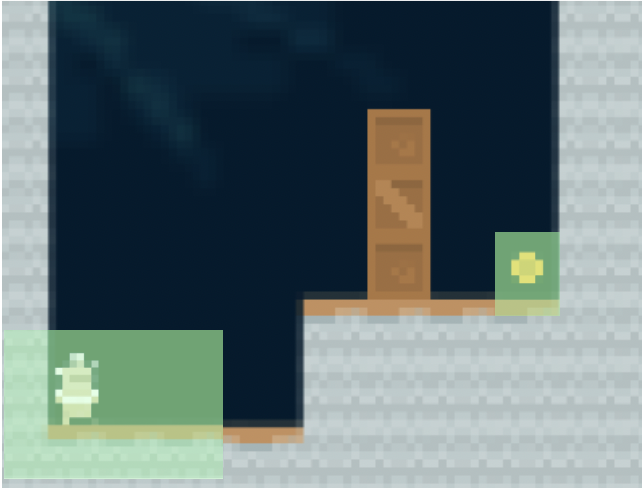}
    \hfill
    \includegraphics[width=0.58\linewidth]{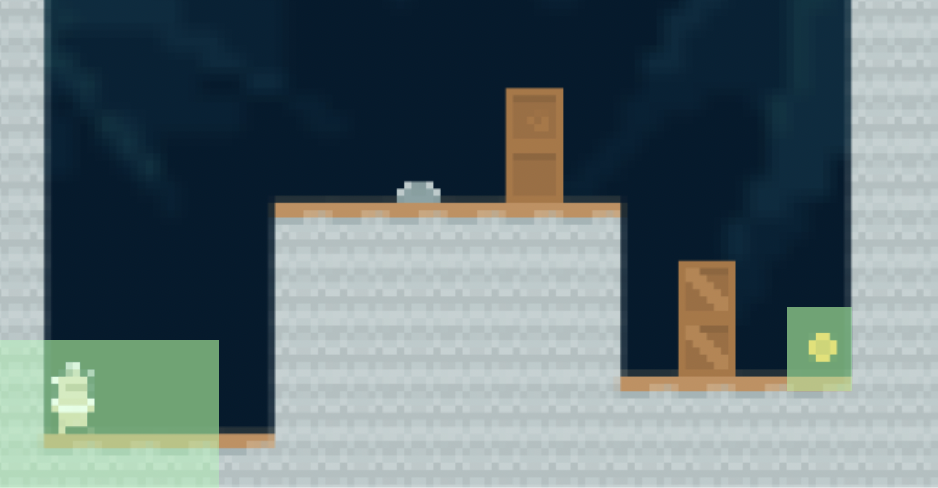}
    % \hfill
    % \includegraphics[width=0.34\linewidth]{Images/coinrun_ex_3.png}
    \caption{A demonstration of the importance of removing unnecessary information from observations for the purpose of policy generalization. In this task, the agent needs to collect the coin at the far right. Both states look very different, but share the same immediate policy (move right). This decision can be reduced from considering the entire observation to just some areas highlighted in green (around the agent, coin location, and position to the right of the agent). In this reduction, the states are almost identical.}
    %}in which case, these states are identical.}
    \label{fig:coinrun_example}
\end{figure}

In order to understand the impact of information removal on policy generalization, we consider environments which have sparse-dependent policies. We consider sparse-dependent policies to be policies in which the next action can be determined based on a small subset of the state variables. This subset is dynamic and can change depending on the state. A policy can be considered sparse-dependent if 
$\pi(s_t) = \pi(\phi(s_t)),$ % \,, \qquad \text{where} \qquad 
where $\phi(s_t) \subset s_t$. % \,.$
Conceptually, a task has a sparse-dependent policy if in most cases the number of state variables needed to compute the next action is much less than the total number of state variables. Note that how many and which of the state variables are useful to the policy is often state-dependent and can be dynamic during an episode, limiting the effectiveness of static feature selection strategies. Based on this idea, we aim to reduce the number of input variables that influence the output in order to learn a policy which better generalizes to unseen states. A task with a sparse dependent policy is illustrated in Figure \ref{fig:coinrun_example}. Environments with sparse dependent policies are quite common and can be considered to lie on a spectrum of policy sparsity. On one end would be very dense maze-like environments, where the agent must consider almost all of the input in order to decide its next action. On the other end would be very sparse reactive environments, where the agent only needs to consider its immediate surroundings to make good choices. Most environments lay somewhere in between and, informally, we find that sparse policies are effective in a wide range of environments. 

\paragraph{Reliable Information Removal}
Many methods aim to ignore portions of the input or otherwise reduce the state representation for the purposes of generalization (discussed further in Section \ref{related_works}). However, usually the behavior of these functions that produce the reduced representations is largely unknown for unseen states. For example, in representation learning methods, the policy is of the form $\pi(\phi(s))$, where $\phi$ is typically some neural network with unknown generalization behavior. Similarly for input masking approaches, which is the focus of comparison in this paper, the policy is of the form $\pi(M(s)\odot s)$, where the masking function $M(s)$ is also some unstructured neural network with unknown generalization ability and $\odot$ denotes the point-wise product. So while this reduced state representation is found reliably for states seen during training, the generalization of the resulting learned policy relies entirely on the ability of the network to extract the correct state representation on unseen states. This presents a significant problem for policy generalization when using the typical unstructured neural networks $M(s)$ for information reduction functions.
%Using the same techniques to improve generalization of $M$ is impractical: 
%sparsifying the input to the mask function $M(s)$ would require the computation of another mask, and so on ad infinitum.

In this paper, we propose a solution to this dilemma: to train a sparse policy
by learning to mask unnecessary attention weights in a self-attention architecture \citep{Vaswani2017}.
We ensure the learned policy is sparse by minimizing the path matrix \citep{spartan},
which measures the combined masking of all possible pathways through the attention network.
As our attention mask is computed based on latent features in each layer of the policy,
sparsity in earlier layers also induces sparse mask computation in later layers.
%(see in Section \ref{mask_gen_properties} for more). 
We hypothesize, and demonstrate empirically, 
that that this mask computation generalizes significantly better when applied to unseen states.

\paragraph{Contributions}
Overall, the contributions of this work are as follows:
(1.) Identify the unreliability associated with many information removal methods for zero-shot policy transfer in multi-task RL.
(2.) Introduce a masked relational policy network which learns to ignore unnecessary relations between state variables, resulting in policies which generalize to more unseen states. 
(3.) Conduct experiments which demonstrate a significant generalization improvement when using sparse relational policies in tasks from the Procgen benchmark.
(4.) A visualization method which shows the importance of each state variables, allowing for improved explainability of the learned policies.

\section{Background}
\label{background}

\subsection{Generalization in RL}
In order to evaluate policy generalization to unseen tasks, we define a set of MDPs $\mathcal{M}$ where each $m_c\in\mathcal{M}$ is one task (also referred to as an instance, level, or context) within this distribution of tasks. Here, each task $m_c$ is the tuple $(\mathcal{S}, \mathcal{A}, T, R, \rho_c, \gamma)$, where $\mathcal{S}$ is the set of states, $\mathcal{A}$ is the set of actions, $T(s'|s,a)$ is the transition probability, $R(s,a)$ is the reward function, $\gamma$ is the discount factor, and $\rho_c(s_0)$ is the starting state distribution.
We assume there is some unknown underlying structure common across all tasks which an agent can use to generalize to unseen tasks $c$. An agent is given a subset of these environments $\mathcal{M}_\text{train}\subset \mathcal{M}$ in which it trains one common policy $\pi$ for all tasks. The generalization of this policy is then evaluated on its performance in the remaining $\mathcal{M}_\text{test} =\mathcal{M}\setminus\mathcal{M}_\text{train}$ tasks. 

For our evaluation, we are using the Procgen benchmark \cite{procgen} which adheres to this formulation. The Procgen benchmark is a set of 16 procedurally generated environments designed to test policy generalization of RL agents. Each instance or level of these environments may differ visually in the coloring and backgrounds or structurally in the positioning of level features like obstacles and enemies. However, the transition dynamics and rewards are shared across all contexts. We follow the protocol originally introduced by \citet{procgen} which has since been frequently used for evaluating policy generalization: per environment, an agent trains on 200 unique levels or contexts~$c$ which determine the MDP $m_c\in\mathcal{M}$. The learned policy is then evaluated on the remaining tasks. 

\subsection{Input Masking}
\label{input-masking}
Information removal methods have been shown to reduce policy reliance on spurious correlations and increase generalization performance to unseen tasks in RL. These methods aim to learn functions which produce an abstraction of the state by removing some amount of unneeded information. This is commonly achieved through data augmentation \cite{drac, soda}, auxiliary losses \cite{dribo, idaac, ctrl}, and input masking \cite{bertoin2022look, grooten2023madi}. While our arguments hold for many forms of information removal, we focus on the input masking approach as it is the most similar to our method and offers the clearest understanding and comparison. In image-based input masking, a function is learned which determines a mask applied to each pixel in the image. In our analysis and comparisons, we use the same setup as in MaDi \cite{grooten2023madi}, where the masking function $\hat M(s)$ is a convolutional network which produces a soft mask with values between 0 and 1 for each pixel. The input image is then multiplied by this mask to produce the reduced information image, which is then sent to the policy function: $\pi\big(\hat M(s)\odot s\big)$. While this method has been found to improve policy generalization to unseen tasks, we note that the generalization properties of the masking function $\hat M$ itself are rather weak. Because $\hat M$ depends on the entire state space, there are no known invariances. We compare our improved masking function generalizability later in Section \ref{mask_gen_properties}.

\subsection{Attention-Based Relational RL}
\label{relational}
The use of relational policies is essential to our method as it allows for the separation of the masking function into individual functions with increased invariance. In relational reinforcement learning \cite{zambaldi2018relational}, the graph relational inductive bias of attention networks is used in order to improve the efficiency, generalization, and interpretability of model-free RL solutions. To achieve this, they use 2D positional encodings and self-attention to create a policy which relates state inputs or entities spatially within the observation. They demonstrate an improvement in sample efficiency and generalization to more complex problems. In order to adapt this architecture for complex image-based inputs, tokenization must be performed. In our method, we use a standard CNN feature extractor, as supported by \citet{sokar2024don} who find that simple tokenization of image inputs in this way is actually beneficial for the performance of online RL algorithms. 

\begin{figure}
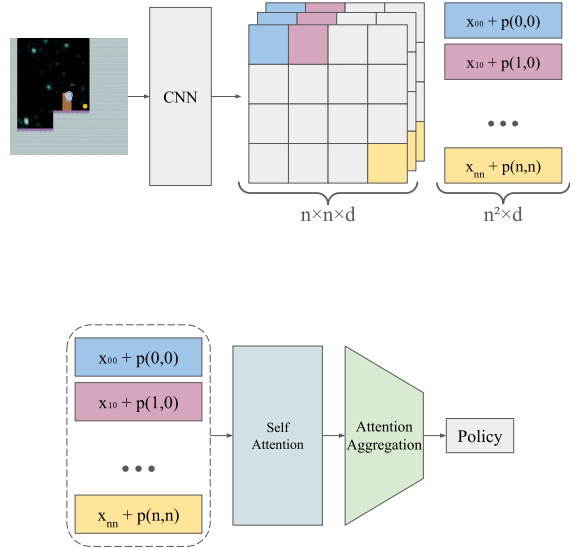

    \centering
    \includesvg[width=0.9\linewidth]{Images/extractor_diagram.svg}
    \hfill
    \includesvg[width=0.9\linewidth]{Images/att_diagram.svg}
    \caption{The CNN feature extractor and positional encoding setup for processing inputs. This set of features are the input tokens to the attention block.}
    \label{fig:extractor}
\end{figure}

\subsection{Learnable Attention Weight Sparsity}
\label{sparse-attention}
The next key component of our method is the sparsification of the masking function, which then induces sparsity in the policy. Our work adapts the attention sparsification methods by \citet{spartan}, who use sparse attention networks in order to learn causal world models and interventions. In order to create sparsity in the attention weights and find a causal structure, they present a stochastic hard attention masking function. This function samples a binary mask $M^l \in \mathbb R^{n \times n}$ for the $n^2$ attention weights in layer $l$ according to a Bernoulli distribution based on the sigmoid $\sigma$ of the unnormalized attention scores $Q^l K^{l\top}$: 
$$
    M^l_{ij} \sim \text{Bern} \big( \sigma \!\left( Q^l{K^{l\top}} \right)_{ij} \big)\,.
$$ 
This binary mask is applied to the attention scores before softmax normalization, giving an output at layer $l$ of 
$$
    h^l = \left( Z^l \oslash (Z^l \mathbbm 1 \mathbbm 1^\top) \right)V^l\,,
    \quad Z^l =  M^l \odot \exp\left({\textstyle\frac{{Q^lK^l}^\top}{\sqrt{d_k}}}\right) \,,
$$ 
where $\oslash$ denotes the point-wise division.
If there is more than one attention layer, it becomes possible for an input node to affect an output node indirectly via other nodes in intermediate layers. In order to measure the influence of each input node on each output node through the multiple attention layers, they additionally define a path matrix 
$$\bar{A} = \left( M^{L} + \mathbb{I} \right) ... \left( M^{2} + \mathbb{I} \right) \left( M^{1} + \mathbb{I} \right) \,,$$
where $\bar{A}_{ij}$, which can be thought of as a kind of adjacency matrix, is the total number of paths through the attention layers connecting input $i$ to output $j$, including the residual connections at each attention layer. \citet{spartan} apply the loss $\mathcal{L}=|\bar{A}|$ to these values in order to reduce the total number of paths and produce a sparse attention network.

%\section{Masked Relational Policies}
\section{Sparse Masked Attention Policies}
\label{methods}
The main contribution of this paper is a novel alternative to input masking for policies,
which addresses the issue of masking functions that do not generalize themselves well.
Instead of directly masking the input, our method extends 
a self-attention architecture described in Section \ref{policy-architecture}.
Sparsity is enforced by drawing binary attention masks in each network layer (Section \ref{relational-masking}).
To make sure the network output is sparse w.r.t.\ the input tokens (Section \ref{relational-paths}),
we minimize a regularization loss that allows us to specify 
the intended sparsity level directly (Section \ref{relational-reduction}).
Finally, we argue why this method generalizes better in Section \ref{mask_gen_properties}.

\subsection{Policy Architecture}
\label{policy-architecture}
Our sparse masked attention policy is built upon the relational architecture used by \citet{zambaldi2018relational} introduced in Section \ref{relational}. We extract visual features from the state and add sinusoidal positional encodings according to their location within the observation. This state tokenizer encoder structure is illustrated in Figure \ref{fig:extractor}. These feature+position tokens are then passed to the self-attention layers, followed by an attention aggregation layer to consolidate the $n$ output nodes, followed finally by a policy head. This architecture results in a policy which is a function of the CNN extracted features, their locations in the observation, and their relations to other extracted feature-position pairs.

\subsection{Relation Masking}
\label{relational-masking}
In order to perform relation masking, we adapt the sparse attention approach from \citet[discussed in Section \ref{sparse-attention}]{spartan}. This allows for dynamic inclusion or exclusion of attention weights. In difference to that work, however, we define for each attention layer $l$ the additional mask embedding functions $Q_m^l$ and $K_m^l$ (distinct from the attention embedding functions $Q^l$ and $K^l$) to determine the probability of masking relations between each pair of input variables. The binary mask $M_{ij}^l$ for each unnormalized attention score $(Q^l{K^{l\top}})_{ij}$ between inputs $i$ and $j$ in attention layer $l$ is sampled from the distribution 
$$
    M^l_{ij} \sim \text{Bern}\Big(\sigma\big((Q_m^l K_m^{l\top})_{ij} + \beta\big)\Big) \,,
$$ 
where $\sigma$ is the sigmoid function and $\beta$ is a bias which affects the probability of a weight being masked upon initialization. The sampling step in the forward pass is performed using the hard Gumbel-Softmax trick to maintain differentiability \cite{gumbelsoftmax, hardconcrete}. In contrast to \citet{spartan}, we must consolidate the $n$ attention outputs in order to produce one policy value. We use an attention aggregation method which provides one output node and computes masked attention between the output nodes at the final layer $L$ and the single final output node. % of dimension $d$.

\subsection{Relational Paths}
\label{relational-paths}
We are interested in the total number of active relations between pairs of input variables. Conceptually, the relation between features $x_i$ and $x_j$ is used by the policy if there are a non-zero number of \emph{paths} through {\em all} attention layers connecting input $x_i$ to output $x_j$ (and vice-versa). We use the same path matrix calculation as \citet{spartan} to calculate the number of paths between pairs of nodes as 
$$A^l = (M^{l-1} + {I_n}) A^{l-1},$$ 
where $A^0 = M^0$. This results in a matrix $A^L \in \mathbb{R}^{n\times n}$ where $A^L_{ij}$ gives the number of unmasked paths between input $x_i$ and $x_j$. Because of our additional aggregation layer, we also compute $A^\text{out} = M^\text{out}A^L$ where $A^\text{out}\in \mathbb{R}^{n}$ and $|A^\text{out}|$ gives the total number of active paths in the network. A small value here indicates a sparse network, while a high value indicates a dense network.

\subsection{Relation Reduction}
\label{relational-reduction}
In order to encourage a sparse relational structure and reduce the number of inputs considered at each node, we apply an additional loss to the policy network which defines the ideal fraction of total paths which remain unmasked, $\alpha \in [0,1]$. At the minimum $\alpha=0$, all relations are masked and the policy uses no information from the input at all. At the maximum $\alpha=1$, the policy uses every attention weight to compute the output, equivalent to using no masking at all:  %. We compute 
$$
    \mathcal{L}_\text{mask} = \mathbb{E}\Big[\big(\alpha - {\textstyle\frac{1}{\mu}} |A^\text{out}| \big)^2\Big] \,,
$$ 
where $\mu$ is the maximum possible total number of active paths through the network. 
Note that a low sparsity (high $\alpha$) does not restrict the ability of the network function class in any way.
Small $\alpha$ {\em enforce} focused attention, though, which can massively improve generalization to unseen states and tasks by removing unnecessary information.%

\begin{figure*}[t!]
    \centering
    \includegraphics[width=0.95\linewidth]{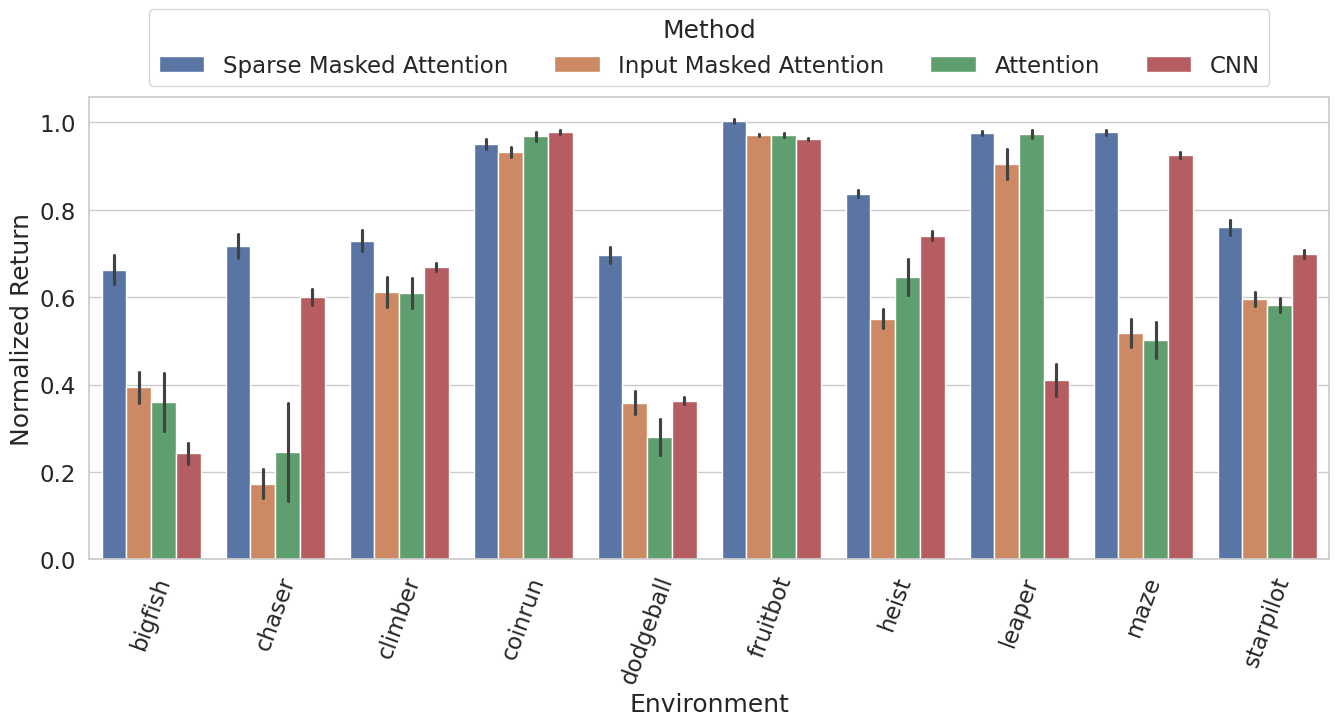}
    \vspace{-1mm}
    \caption{The normalized returns in unseen tasks for each of the Procgen games. %The agents were all trained for 25M timesteps on 200 contexts and evaluated on 100 contexts randomly sampled from the remaining unused contexts.
    The mean and standard error over 10 seeds is plotted.}
    \label{fig:procgen-full-bar}
\end{figure*}

\subsection{Generalization Properties of the Masking Function}
\label{mask_gen_properties}
As discussed previously, the generalization of the abstraction/masking function is just as important as the generalization of the policy, but this is frequently ignored. A key benefit of our attention weight masking method (compared to input masking or other information removal methods) is that the masking function is integrated into the sparsified network, yielding strong generalization properties. 

To illustrate this strong generalization ability, consider the functional differences between simple input masking $\hat M$ \citep{bertoin2022look,grooten2023madi} and attention weight masking $M$ \citep{spartan}. When performing input masking, the masking function $\hat M$ computes $n$ masks by operating on the entire state $s = \{x_1, x_2, ..., x_n\}$. This gives a masking function of the form
$$
    \hat M(x_1, x_2, \ldots, x_n) \in [0,1]^{n}\,,
$$ 
where the $i$th entry $\hat M_i(x_1, x_2, ..., x_n)$ can be either a soft mask value or the probability of masking out input $x_i$.
\\
In contrast, when performing attention weight masking, the masking function $M$ computes $n^2$ masks by operating on all pairs of input variables, ignoring the other $n-2$ variables in each mask calculation:
$$
    M(x_i,x_j) \in [0,1] \,, \quad \forall i,j \,.
$$
While this masks out individual relations, entire input variables $x_i$ are masked out when no $x_j$ pays attention:
$$
    \hat M_i(x_1, \ldots, x_n) \;\equiv\; {\textstyle\prod_j} \, M(x_i, x_j) \,.
$$ 
This clearly restricts the model class of the effective input mask in the first attention layer.
But like general computation in attention networks, stacking multiple layers can compensate for that. 
From the second layer on each attention mask is computed based on the output tokens of the previous layer, which can depend on arbitrary many (or few) input variables. 
Enforcing sparsity in previous layers therefore also yields sparse mask computation in later layers.
We can still efficiently compute the sparsity loss of the corresponding input mask, the product of all possible paths though the network, with the path matrix $A^\text{out}$.% \citep[see Section \ref{sparse-attention},][]{spartan}.

Mask computation generalizes thus due to architectural restrictions in early layers, and sparsification in later layers.

\section{Experiments}
\label{experiments}
We conduct experiments to evaluate and analyze our method using environments from the Procgen \cite{procgen} benchmark. Procgen contains a diverse set of environments with complex visual inputs and procedurally generated tasks which allow for large numbers of train and test contexts. We conduct experiments to evaluate our method's ability to produce a policy which generalizes to unseen tasks compared to relevant baselines. Additionally, we conduct experiments to evaluate how different levels of sparsity affect the generalization ability. Finally, we take advantage of the attention structure to visualize and discuss the state variable dependencies learned by dense and sparse policies and show that these features tend to be more generalizable. 

\subsection{Experimental setup and methods}
The baseline methods we compare with are chosen to examine the different aspects of information removal and how they affect generalization to unseen tasks. All of our baselines use the same PPO hyperparameters and CNN feature extractor to provide the closest comparison. For all methods and environments, agents are given 200 tasks to train on and then are evaluated on 100 (tasks) contexts randomly sampled from the remaining unused contexts. % unseen tasks. 
Policies are trained with PPO for 25 million timesteps. Further implementation details are provided in the appendix.\\
\label{baselines}%
\textbf{Baseline: CNN Policy} A standard PPO setup \citep{procgen}: a base CNN feature extractor is followed by a MLP. \\
\textbf{Baseline: Attention Policy} This is an attention-based policy network using 2D positional encodings \citep{zambaldi2018relational}. % Each CNN output is given an (x,y) positional encoding according to its location in the feature map. These features are the input tokens to the attention network. 
This architecture is illustrated in Figure \ref{fig:extractor}. \\
\textbf{Baseline: Input Masked Attention} This is an extension of the Attention policy which introduces an additional masking network as in \cite{grooten2023madi}. The learned pixel mask is applied to the observation before being passed to CNN of the the baseline Attention policy. \\
\textbf{Contribution: Sparse Masked Attention Policy} This is our extension of the attention policy where we add an additional masking function which applies a binary mask to each attention weight in the network, as described in Section \ref{relational-masking}. The additional loss of Section \ref{relational-reduction} is applied (unless otherwise stated with $\alpha=0.05$), which specifies the fraction of total paths that the network is allowed to use.

\subsection{Procgen Test Performance}
\label{exp-procgen}
Figure \ref{fig:procgen-full-bar} shows the final test performance of our method and the baselines on each of the Procgen environments. The return values are normalized according to approximate minimum and maximum rewards presented in \cite{procgen}. A table with raw, unnormalized returns for each environment can be found in the appendix in Table \ref{tab:procgen-test}.

First, we see that our Sparse Masked Attention (blue) networks 
significantly outperform their dense counterparts (green) in most environments,
and performs at least equal in environments (\textit{fruitbot, coinrun, leaper}) 
where policies do not seem to benefit from sparsity.
%consistently outperform their dense counterparts (green), 
This demonstrates the importance of learning sparse policies for generalization to unseen tasks.

Second, we see that while input masking (yellow) can improve slightly over the unmasked attention, this benefit is inconsistent and occasionally results even in worse performance in unseen states. We argue that this is due to the difficulty of learning a state-dependent mask that generalizes to unseen inputs, which our method improves upon.

Lastly, the (red) CNN baseline is sometimes surprisingly strong (e.g., \textit{chaser}), 
but also extremely inconsistent (e.g., \textit{leaper}).
Tuning hyper-parameters to individual environments could make attention and CNN baselines more comparable.
However, we interpret the clear advantage of sparse attention in most environments 
as evidence for the robustness of our method, making such fine tuning unnecessary.  

\begin{figure}[t!]
    \centering
    \includegraphics[width=1\linewidth]{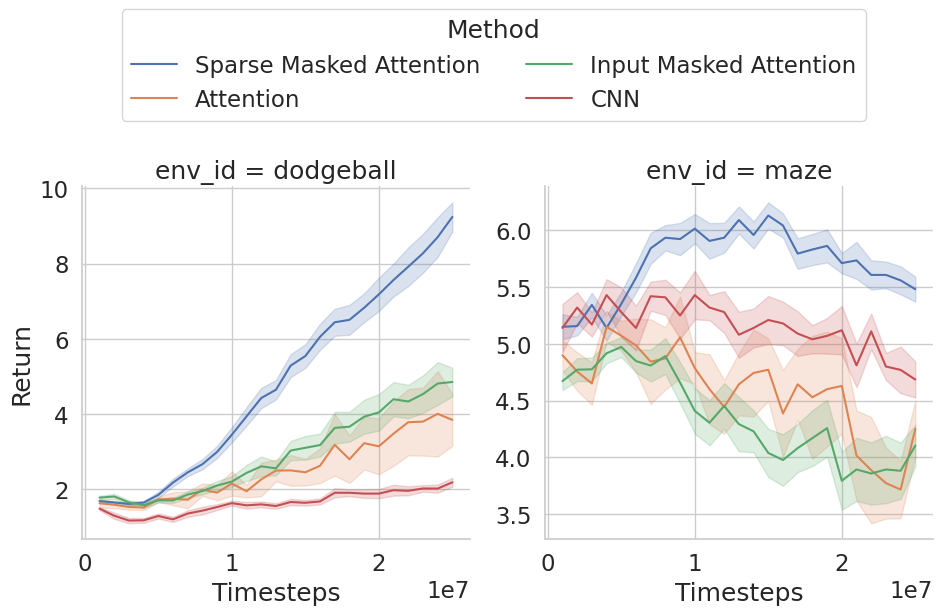}
    \caption{The test returns (mean and standard error over 10 seeds) on unseen tasks from the \textit{dodgeball} and \textit{maze} environments. As expected, sparse masking results in a large improvement in generalization for dodgeball, which has an underlying sparse dependent policy structure, but shows much smaller improvement in maze, whose underlying policy structure is relatively dense.}
    \label{fig:procgen_sample_test}
\end{figure}

\subsection{The Impact of Sparsity on Generalization}

% \begin{figure*}
%     \centering
%     \includegraphics[width=0.95\linewidth]{Images/procgen_sample_sparsity.png}
%     \caption{The test performance for different sparsity levels.}
%     \label{fig:procgen-sparsity}
% \end{figure*}

\begin{figure}[t!]
    \centering
    \includegraphics[width=1\linewidth]{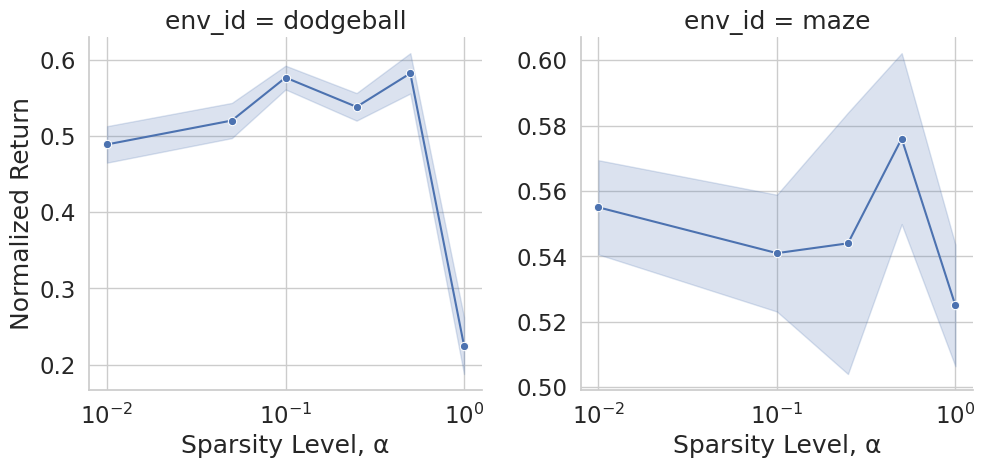}
    \caption{The normalized test return (mean and standard error over 10 seeds)
            of our method for different sparsity levels $\alpha$
            in the \textit{dodgeball} and \textit{maze} environments.
            For more environments see Figure \ref{fig:all-sparsitiy-levels} in the appendix.}
    \label{fig:procgen-sparsity}
\end{figure}

\begin{figure*}[tb]
    \centering
    \begin{tabular}{>{\centering\arraybackslash}m{0.1\linewidth} >{\centering\arraybackslash}m{0.8\linewidth}}
    Train Task & \includegraphics[width=0.9\linewidth]{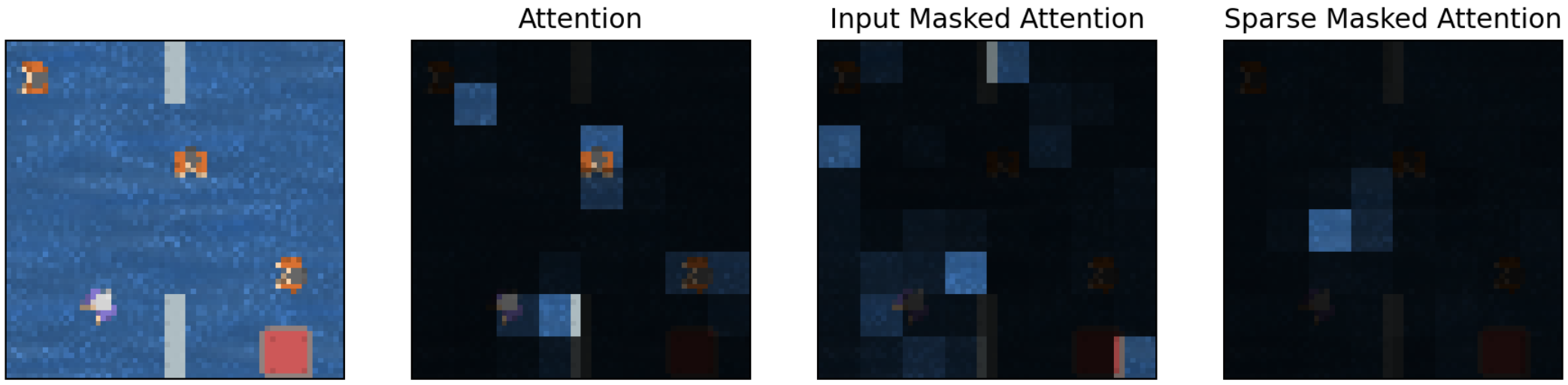} \\
    Unseen Test Task & \includegraphics[width=0.9\linewidth]{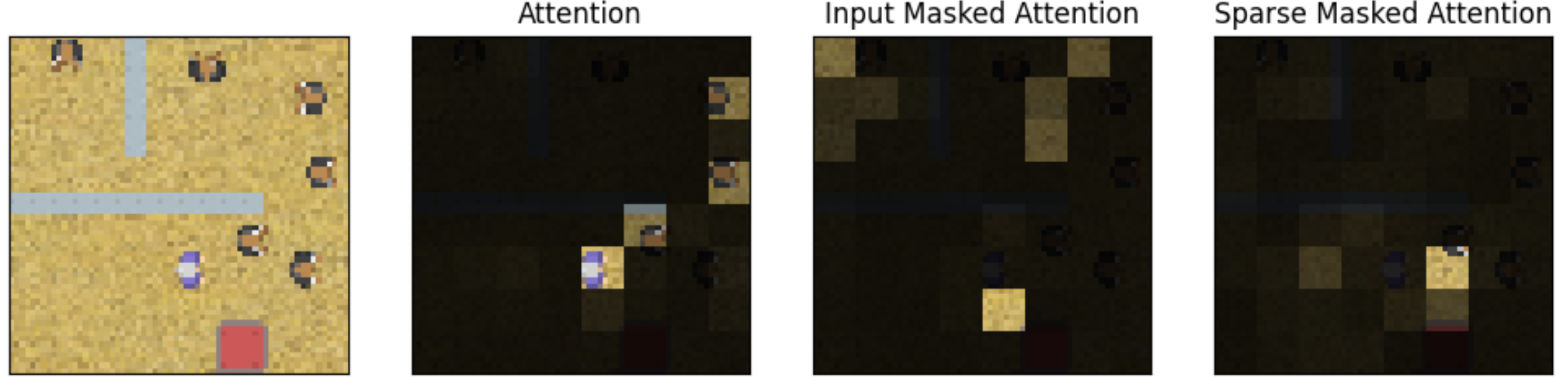}\\
    \end{tabular}
    \caption{The relative importance of each area of the observation according to the attention maps of policies learned from Attention, Input Masked Attention, and Sparse Masked Attention networks on the Dodgeball environment. }
    \label{fig:viz-dodgeball}
    \vspace{-1mm}
\end{figure*}

\label{sparsity_analysis}
Above we demonstrate significant improvements in generalization performance when using our method, which we hypothesize is a result of increased sparsity in both the policy and the masking function. 
Not all environments admit {\em sparse dependent policies} (see the discussion in Section~\ref{introduction}),
so we see strong differences between environments. To further investigate
the effect of different sparsity levels on generalization performance, 
we chose the \textit{dodgeball} and \textit{maze} environments as two opposite examples. 
Figure \ref{fig:procgen_sample_test} shows the development of test returns from these environments 
throughout training.
Note that our proposed method improves generalization a lot in \textit{dodgeball}, but only a little in \textit{maze}.
We expect that a target fraction of nonzero paths of $\alpha=1$ should achieve similar performance to the unmasked relational policy. 
Decreasing this fraction (and therefore increasing sparsity) in environments that admit sparse dependent policies should improve test performance, but only up to a point. 
At some limit more sparsity will no longer benefit generalization, 
as the mask will have removed too much information 
and the policy must again rely on overfitting to small, 
specific details which are unlikely to be task-invariant. 
After this point, we expect generalization performance to fall. 

In Figure \ref{fig:procgen-sparsity}, we show the test performance of our sparse masked attention policy on unseen tasks at different levels of sparsity enforced 
by the loss from Section \ref{relational-reduction}.
%within the mask for the dodegball and maze environments. 
As expected, in \textit{dodgeball} we see that any level of sparsity improves generalization, highlighting the importance of removing overfitting information. This aligns with our understanding that in a sparse, reactive environments, policy sparsity can result in large improvements to generalization, as there is a lot of unnecessary information in the observation. In the \textit{maze} environment, we see that sparsity offers much less benefit to generalization. This also aligns with our understanding of sparse-dependent policies. Because a maze requires knowledge of almost all locations within the maze, there is not much sparsity that can be enforced. Additionally, we see that there is a wide range of sparsity levels that will produce well-generalizing policies, indicating that our method is robust to this choice of hyperparameter.
Similar effects can be observed in all tested Procgen environments, which can be found in Figure \ref{fig:all-sparsitiy-levels} of the appendix.

\subsection{Attention Map Visualizations}
\label{visualizations}
\begin{figure*}[tb]
    \centering
    \begin{tabular}{>{\centering\arraybackslash}m{0.1\linewidth} >{\centering\arraybackslash}m{0.8\linewidth}}
    Train Task & \includegraphics[width=0.9\linewidth]{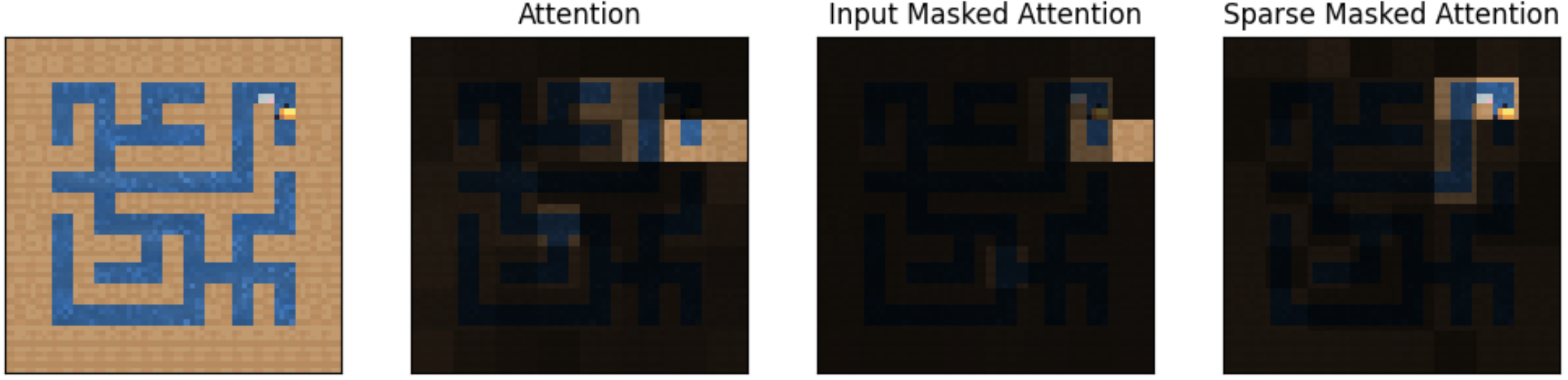} \\
    Unseen Test Task & \includegraphics[width=0.9\linewidth]{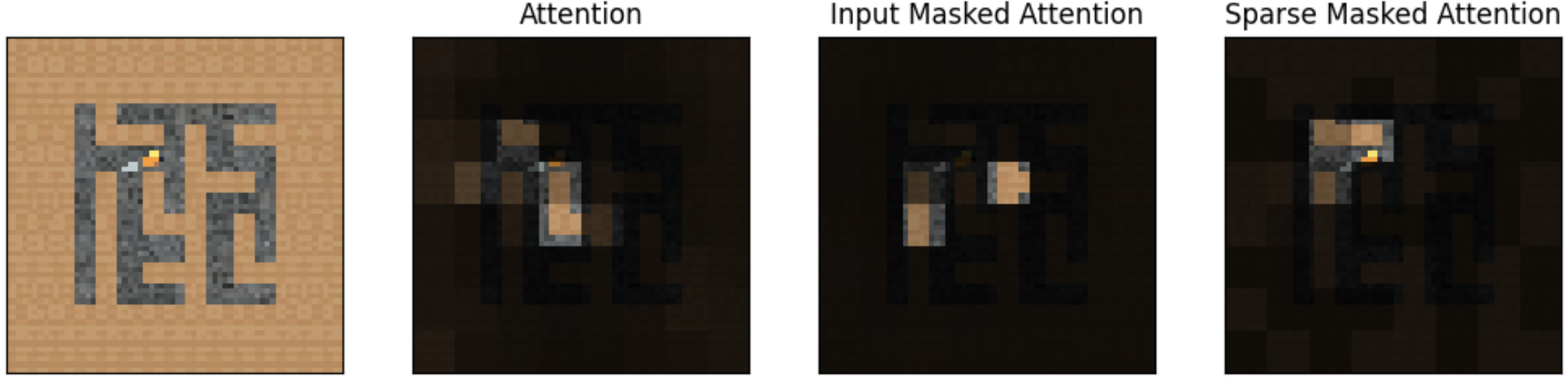} \\
    \end{tabular}
    \caption{The relative importance of each area of the observation according to the attention maps of policies learned from Attention, Input Masked Attention, and Sparse Masked Attention networks on the Maze environment.}
    \label{fig:viz-maze}
\end{figure*}

A nice feature of attention networks is the ability to examine the total attention paid to each input. We can consider this %total attention paid to each input 
as a measure of importance of that input for the policy. In Figures \ref{fig:viz-dodgeball} and \ref{fig:viz-maze}, we present an example of an observation and the attention paid to each area of that observation by Attention, Input Masked Attention, and Sparse Masked Attention policies on the Dodgeball and Maze environments. 
For the dodgeball task, the agent (purple) must avoid walls, enemies, and any dodgeballs thrown at them. We see that in the training task, the Attention model focuses on the agent, enemies, and some structural features. Most of this information is not needed and will not transfer well to new tasks. The Input Masked Attention model seems to successfully focus on the area around the agent and ignore the far away enemies, but still relies on some structural features. This also indicates some level of overfitting where the structural features can be used to determine the level. In our Sparse Masked Attention model, we see that the agent focuses only on the surroundings of the agent, which is a well-generalizing abstraction. To evaluate how well these learned abstraction functions operate in unseen environments, we look at an observation from the unseen test task. We observe that the Attention model again focuses on the agent and multiple enemies. While this is a successful generalization of the abstraction, this is not a well-generalizing selection of features. Looking at the Input Masked Attention in the unseen task, we see that while the agent does focus on an area around the agent, the other information used by the policy does not appear meaningful and indicates a failure of the masking function to generalize to an unseen state. In our Sparse Masked Attention method, we see that the agent successfully focuses on the areas it is facing. This is both a well-generalizing abstraction for the policy, and it aligns with the masks seen in training, indicating that the masking function has generalized well.

\section{Related Work}
\label{related_works}
% \paragraph{Information Removal}
\paragraph{Sparsity}
Most previous uses of sparsity in reinforcement learning focus on reducing model size and improving computational and sample efficiency: \citet{painter2012greedy} consider various sparse feature selection algorithms for value-based RL, \citet{arnob2025sparse} introduce a sparsity regularizer that is applied to the neural networks of various offline RL algorithms, and \citet{sokar2021dynamic} learn dynamic sparse neural networks, all resulting in faster training. In contrast, our method aims to use the properties of attention to find policies which generalize well to unseen data rather than speeding up policy learning in training tasks.
\paragraph{Masking}
 Input masking is a form of information removal and sparsity that has been used to improve generalization to unseen states. \citet{grooten2023madi} and \citet{bertoin2022look} address input sparsity/reduction by learning a dynamic mask for each pixel which multiplies the input image before it is sent to the policy network. This allows for the removal of unnecessary or overfitting information in the observation, but unlike our method, does not produce a reliable masking function for unseen data.
\paragraph{Attention and Transformers}
Attention networks have also been used for improving generalization in RL. \citet{zambaldi2018relational} exploit the relational structure underlying many RL tasks and use a similar spatial-relational structure as our method, demonstrating improved policy generalization, particularly in extrapolation to larger problem sizes. However, unlike our method, this does not attempt to induce sparsity or reduce information in the policy, which leaves room for overfitting. \citet{weissenbacher2024sit} use graph-symmetric attention in order to encode local and global visual symmetries and invariances into their network, \citet{sensory-neuron} use attention networks to demonstrate how permutation invariance affects generalization and robustness in RL, and \citet{att-agent} use neuroevolution to find a self-attention based importance voting system for input image patches, resulting in agents which are better at ignoring distractions in the input. 

\section{Conclusion}
\label{conclusion}
In this paper, we presented a new approach to improving policy generalization to unseen tasks in deep reinforcement learning. We identified that previous methods are able to improve generalization by performing information removal, however, they fail to consider the generalization properties of the information removal method itself, limiting its effectiveness. We proposed to address this problem by using an information removal method which has known strong generalization properties in unseen observations. We presented our method of Sparse Masked Attention policies which incorporates a learned attention weight masking function into an attention-based policy network. This form of masking produces a sparse, well-generalizing policy, but also performs more reliable information removal due to the sparsity and attention structure of the masking function itself. We evaluated our method in the Procgen benchmark and demonstrated that our method improved policy generalization to unseen tasks when compared to dense, unmasked attention networks, and attention networks with input masking functions. 
%We also observed a direct relationship between increased sparsity and increased generalization performance. As we hypothesized, this relationship is more pronounced in environments which have an underlying sparse-dependent policy structure. 
In most environments we observed that a sufficient level of sparsity increased generalization performance, sometimes dramatically. In other environments, which we hypothesize do not allow for a sparse-dependent policy structure, sparsity levels did not seem to affect the performance much. This indicates a wide applicability of our method, even in environments not perfectly suited.
When comparing the features used by each policy, we found that our Sparse Masked Attention policies more reliably identified generalizable features within the observation which we believe to be a result of our well-generalizing masking function. 

There are some assumptions and limitations to be aware of regarding this method. One limitation is the allowed relations hyperparameter $\alpha$. While this parameter is not very sensitive and a very wide range of parameters provide generalization improvement, it is not known what the ideal ratio is here or how this may be efficiently optimized. Another possible limitation is in the structure of the observation. We assume that features are somewhat localized and separated within the observation. While this is commonly true in many benchmarks, it should be noted that the CNN feature extraction and positional encodings may not be effective for tasks which rely on global or entangled features.  

\section{Acknowledgments}
We would like to thank Ariel Ebersberger for discussions and insights relating to this work. This work was partially funded by the Dutch Research Council (NWO) project Reliable Out-of-Distribution Generalization in Deep Reinforcement Learning with project number OCENW.M.21.234. Research reported in this work was partially or completely facilitated by computational resources and support of the \citet{DAIC} at TU Delft and the DelftBlue supercomputer provided by the \citet{DHPC2024}.

\section*{Impact Statement}
This paper presents work whose goal is to advance the field of Machine
Learning. There are many potential societal consequences of our work, none
which we feel must be specifically highlighted here.

\bibliography{references}

\begin{thebibliography}{23}
\providecommand{\natexlab}[1]{#1}
\providecommand{\url}[1]{\texttt{#1}}
\expandafter\ifx\csname urlstyle\endcsname\relax
  \providecommand{\doi}[1]{doi: #1}\else
  \providecommand{\doi}{doi: \begingroup \urlstyle{rm}\Url}\fi

\bibitem[Arnob et~al.(2025)Arnob, Fujimoto, and Precup]{arnob2025sparse}
Arnob, S.~Y., Fujimoto, S., and Precup, D.
\newblock Sparse-reg: Improving sample complexity in offline reinforcement learning using sparsity.
\newblock \emph{arXiv preprint arXiv:2506.17155}, 2025.

\bibitem[Bertoin et~al.(2022)Bertoin, Zouitine, Zouitine, and Rachelson]{bertoin2022look}
Bertoin, D., Zouitine, A., Zouitine, M., and Rachelson, E.
\newblock Look where you look! saliency-guided q-networks for generalization in visual reinforcement learning.
\newblock \emph{Advances in neural information processing systems}, 35:\penalty0 30693--30706, 2022.

\bibitem[Cobbe et~al.(2020)Cobbe, Hesse, Hilton, and Schulman]{procgen}
Cobbe, K., Hesse, C., Hilton, J., and Schulman, J.
\newblock Leveraging procedural generation to benchmark reinforcement learning, 2020.
\newblock URL \url{https://arxiv.org/abs/1912.01588}.

\bibitem[(DAIC)(2024)]{DAIC}
(DAIC), D. A.~C.
\newblock The delft ai cluster (daic), rrid:scr\_025091, 2024.
\newblock URL \url{https://doc.daic.tudelft.nl/}.

\bibitem[{D}elft {H}igh {P}erformance {C}omputing~{C}entre ({DHPC})(2024)]{DHPC2024}
{D}elft {H}igh {P}erformance {C}omputing~{C}entre ({DHPC}).
\newblock \emph{{D}elft{B}lue {S}upercomputer ({P}hase 2)}, 2024.
\newblock \url{https://www.tudelft.nl/dhpc/ark:/44463/DelftBluePhase2}.

\bibitem[Fan \& Li(2022)Fan and Li]{dribo}
Fan, J. and Li, W.
\newblock Dribo: Robust deep reinforcement learning via multi-view information bottleneck, 2022.
\newblock URL \url{https://arxiv.org/abs/2102.13268}.

\bibitem[Grooten et~al.(2023)Grooten, Tomilin, Vasan, Taylor, Mahmood, Fang, Pechenizkiy, and Mocanu]{grooten2023madi}
Grooten, B., Tomilin, T., Vasan, G., Taylor, M.~E., Mahmood, A.~R., Fang, M., Pechenizkiy, M., and Mocanu, D.~C.
\newblock Madi: Learning to mask distractions for generalization in visual deep reinforcement learning.
\newblock \emph{arXiv preprint arXiv:2312.15339}, 2023.

\bibitem[Hansen \& Wang(2021)Hansen and Wang]{soda}
Hansen, N. and Wang, X.
\newblock Generalization in reinforcement learning by soft data augmentation, 2021.
\newblock URL \url{https://arxiv.org/abs/2011.13389}.

\bibitem[Jang et~al.(2017)Jang, Gu, and Poole]{gumbelsoftmax}
Jang, E., Gu, S., and Poole, B.
\newblock Categorical reparameterization with gumbel-softmax, 2017.
\newblock URL \url{https://arxiv.org/abs/1611.01144}.

\bibitem[Kirk et~al.(2023)Kirk, Zhang, Grefenstette, and Rockt{\"a}schel]{kirk2023survey}
Kirk, R., Zhang, A., Grefenstette, E., and Rockt{\"a}schel, T.
\newblock A survey of zero-shot generalisation in deep reinforcement learning.
\newblock \emph{Journal of Artificial Intelligence Research}, 76:\penalty0 201--264, 2023.

\bibitem[Lei et~al.(2025)Lei, Schölkopf, and Posner]{spartan}
Lei, A., Schölkopf, B., and Posner, I.
\newblock {SPARTAN}: A sparse transformer world model attending to what matters, 2025.
\newblock URL \url{https://arxiv.org/abs/2411.06890}.

\bibitem[Maddison et~al.(2017)Maddison, Mnih, and Teh]{hardconcrete}
Maddison, C.~J., Mnih, A., and Teh, Y.~W.
\newblock The concrete distribution: A continuous relaxation of discrete random variables, 2017.
\newblock URL \url{https://arxiv.org/abs/1611.00712}.

\bibitem[Mazoure et~al.(2022)Mazoure, Ahmed, MacAlpine, Hjelm, and Kolobov]{ctrl}
Mazoure, B., Ahmed, A.~M., MacAlpine, P., Hjelm, R.~D., and Kolobov, A.
\newblock Cross-trajectory representation learning for zero-shot generalization in rl, 2022.
\newblock URL \url{https://arxiv.org/abs/2106.02193}.

\bibitem[Painter-Wakefield \& Parr(2012)Painter-Wakefield and Parr]{painter2012greedy}
Painter-Wakefield, C. and Parr, R.
\newblock Greedy algorithms for sparse reinforcement learning.
\newblock \emph{arXiv preprint arXiv:1206.6485}, 2012.

\bibitem[Raileanu \& Fergus(2021)Raileanu and Fergus]{idaac}
Raileanu, R. and Fergus, R.
\newblock Decoupling value and policy for generalization in reinforcement learning, 2021.
\newblock URL \url{https://arxiv.org/abs/2102.10330}.

\bibitem[Raileanu et~al.(2021)Raileanu, Goldstein, Yarats, Kostrikov, and Fergus]{drac}
Raileanu, R., Goldstein, M., Yarats, D., Kostrikov, I., and Fergus, R.
\newblock Automatic data augmentation for generalization in deep reinforcement learning, 2021.
\newblock URL \url{https://arxiv.org/abs/2006.12862}.

\bibitem[Sokar et~al.(2021)Sokar, Mocanu, Mocanu, Pechenizkiy, and Stone]{sokar2021dynamic}
Sokar, G., Mocanu, E., Mocanu, D.~C., Pechenizkiy, M., and Stone, P.
\newblock Dynamic sparse training for deep reinforcement learning.
\newblock \emph{arXiv preprint arXiv:2106.04217}, 2021.

\bibitem[Sokar et~al.(2024)Sokar, Obando-Ceron, Courville, Larochelle, and Castro]{sokar2024don}
Sokar, G., Obando-Ceron, J., Courville, A., Larochelle, H., and Castro, P.~S.
\newblock Don't flatten, tokenize! unlocking the key to softmoe's efficacy in deep rl.
\newblock \emph{arXiv preprint arXiv:2410.01930}, 2024.

\bibitem[Tang \& Ha(2021)Tang and Ha]{sensory-neuron}
Tang, Y. and Ha, D.
\newblock The sensory neuron as a transformer: Permutation-invariant neural networks for reinforcement learning, 2021.
\newblock URL \url{https://arxiv.org/abs/2109.02869}.

\bibitem[Tang et~al.(2020)Tang, Nguyen, and Ha]{att-agent}
Tang, Y., Nguyen, D., and Ha, D.
\newblock Neuroevolution of self-interpretable agents.
\newblock In \emph{Proceedings of the 2020 Genetic and Evolutionary Computation Conference}, GECCO ’20, pp.\  414–424. ACM, June 2020.
\newblock \doi{10.1145/3377930.3389847}.
\newblock URL \url{http://dx.doi.org/10.1145/3377930.3389847}.

\bibitem[Vaswani et~al.(2017)Vaswani, Shazeer, Parmar, Uszkoreit, Jones, Gomez, Kaiser, and Polosukhin]{Vaswani2017}
Vaswani, A., Shazeer, N., Parmar, N., Uszkoreit, J., Jones, L., Gomez, A.~N., Kaiser, {\L}., and Polosukhin, I.
\newblock Attention is all you need.
\newblock In \emph{Advances in neural information processing systems}, pp.\  5998--6008, 2017.

\bibitem[Weissenbacher et~al.(2024)Weissenbacher, Agarwal, and Kawahara]{weissenbacher2024sit}
Weissenbacher, M., Agarwal, R., and Kawahara, Y.
\newblock Sit: symmetry-invariant transformers for generalisation in reinforcement learning.
\newblock \emph{arXiv preprint arXiv:2406.15025}, 2024.

\bibitem[Zambaldi et~al.(2018)Zambaldi, Raposo, Santoro, Bapst, Li, Babuschkin, Tuyls, Reichert, Lillicrap, Lockhart, et~al.]{zambaldi2018relational}
Zambaldi, V., Raposo, D., Santoro, A., Bapst, V., Li, Y., Babuschkin, I., Tuyls, K., Reichert, D., Lillicrap, T., Lockhart, E., et~al.
\newblock Relational deep reinforcement learning.
\newblock \emph{arXiv preprint arXiv:1806.01830}, 2018.

\end{thebibliography}
\bibliographystyle{icml2026}

%%%%%%%%%%%%%%%%%%%%%%%%%%%%%%%%%%%%%%%%%%%%%%%%%%%%%%%%%%%%%%%%%%%%%%%%%%%%%%%
%%%%%%%%%%%%%%%%%%%%%%%%%%%%%%%%%%%%%%%%%%%%%%%%%%%%%%%%%%%%%%%%%%%%%%%%%%%%%%%
% APPENDIX
%%%%%%%%%%%%%%%%%%%%%%%%%%%%%%%%%%%%%%%%%%%%%%%%%%%%%%%%%%%%%%%%%%%%%%%%%%%%%%%
%%%%%%%%%%%%%%%%%%%%%%%%%%%%%%%%%%%%%%%%%%%%%%%%%%%%%%%%%%%%%%%%%%%%%%%%%%%%%%%
\newpage
\appendix
\onecolumn
\section{Appendix}

\begin{table}[h]
    \centering
    \begin{tabular}{l|c|c|c|c}
         Environment & CNN & Attention & Input Masked Attention & \textbf{Sparse Masked Attention (ours)} \\
         \hline
         \hline
         bigfish & 2.35 $\pm$ 0.35 & 11.32 $\pm$ 2.14 & 11.96 $\pm$ 1.16 & \ 21.61 $\pm$ 0.87 \\
        fruitbot & 27.68 $\pm$ 0.28 & 30.07 $\pm$ 0.21 & 29.61 $\pm$ 0.18 & 30.74 $\pm$ 0.20 \\
        coinrun & 8.92 $\pm$ 0.06 & 8.42 $\pm$ 0.15 & 8.21 $\pm$ 0.12 & 8.84 $\pm$ 0.10 \\
        dodgeball & 2.19 $\pm$ 0.12 & 3.85 $\pm$ 0.69 & 4.86 $\pm$ 0.38 & 10.95 $\pm$ 0.30 \\
        heist & 1.98 $\pm$ 0.09 & 2.62 $\pm$ 0.23 & 2.74 $\pm$ 0.16 & 4.12 $\pm$ 0.23 \\
        starpilot & 42.56 $\pm$ 0.57 & 36.69 $\pm$ 1.21 & 38.43 $\pm$ 1.32 & 46.38 $\pm$ 0.89 \\
        climber & 5.01 $\pm$ 0.18 & 5.71 $\pm$ 0.56 & 5.54 $\pm$ 0.31 & 6.76 $\pm$ 0.20 \\
        maze & 4.69 $\pm$ 0.16 & 4.88 $\pm$ 0.29 & 4.49 $\pm$ 0.25 & 5.41 $\pm$ 0.18 \\
        chaser & 6.12 $\pm$ 0.29 & 2.71 $\pm$ 1.16 & 2.25 $\pm$ 0.43 & 8.73 $\pm$ 0.42 \\
        leaper & 3.71 $\pm$ 0.31 & 8.88 $\pm$ 0.07 & 8.39 $\pm$ 0.32 & 9.11 $\pm$ 0.09 \\
        
    \end{tabular}
    \caption{Returns in unseen test environments (mean and standard error over 10 seeds) on various Procgen environments for our method (Sparse Masked Attention) and the baselines we compare against. }
    \label{tab:procgen-test}
\end{table}

\begin{table}[h]
    \centering
    \begin{tabular}{l|c|c|c|c}
         Environment & CNN & Attention & Input Masked Attention & \textbf{Sparse Masked Attention (ours)} \\
         \hline
         \hline
         bigfish & 9.73 $\pm$ 0.97 & 14.42 $\pm$ 2.64 & 15.75 $\pm$ 1.41 & 26.51 $\pm$ 1.31 \\
        fruitbot & 31.15 $\pm$ 0.09 & 31.45 $\pm$ 0.14 & 31.46 $\pm$ 0.10 & 32.49 $\pm$ 0.13 \\
        coinrun & 9.78 $\pm$ 0.04 & 9.68 $\pm$ 0.10 & 9.32 $\pm$ 0.11 & 9.51 $\pm$ 0.11 \\
        dodgeball & 6.90 $\pm$ 0.16 & 5.33 $\pm$ 0.79 & 6.81 $\pm$ 0.51 & 13.24 $\pm$ 0.36 \\
        heist & 7.41 $\pm$ 0.10 & 6.46 $\pm$ 0.42 & 5.51 $\pm$ 0.22 & 8.36 $\pm$ 0.08 \\
        starpilot & 44.68 $\pm$ 0.59 & 37.27 $\pm$ 1.04 & 38.15 $\pm$ 1.03 & 48.61 $\pm$ 1.12 \\
        climber & 8.42 $\pm$ 0.12 & 7.69 $\pm$ 0.43 & 7.71 $\pm$ 0.42 & 9.19 $\pm$ 0.30 \\
        maze & 9.26 $\pm$ 0.06 & 5.78 $\pm$ 0.70 & 6.34 $\pm$ 0.53 & 9.77 $\pm$ 0.05 \\
        chaser & 7.80 $\pm$ 0.25 & 3.19 $\pm$ 1.46 & 2.26 $\pm$ 0.42 & 9.33 $\pm$ 0.36 \\
        leaper & 4.11 $\pm$ 0.36 & 9.74 $\pm$ 0.10 & 9.04 $\pm$ 0.34 & 9.75 $\pm$ 0.05 \\
        
    \end{tabular}
    \caption{Returns in training environments (mean and standard error over 10 seeds) on various Procgen environments for our method (Sparse Masked Attention) and the baselines we compare against. }
\end{table}

\begin{figure}[h]
    \centering
    \includegraphics[width=1\linewidth]{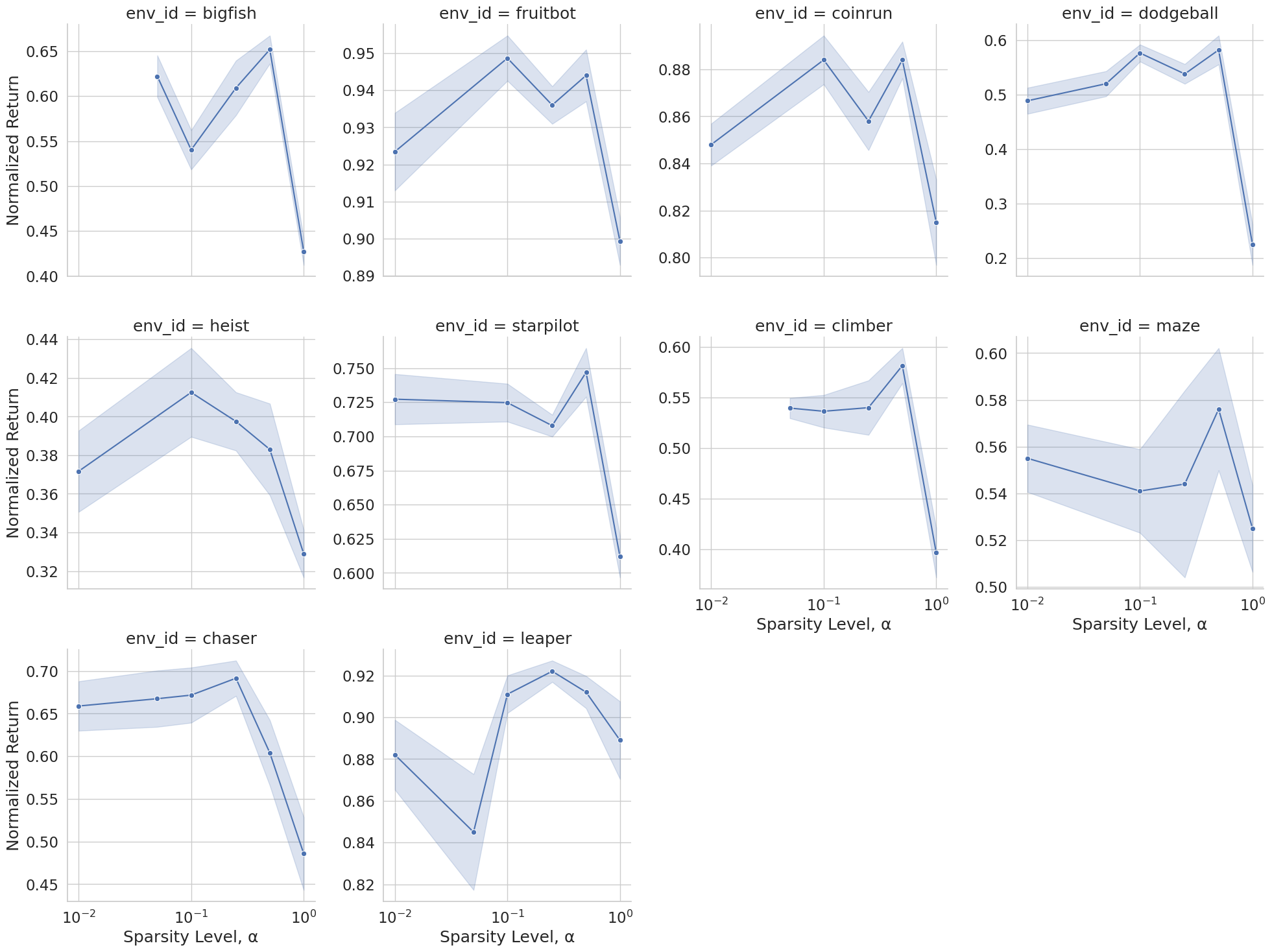}
    \caption{The normalized return (mean and standard error over 10 seeds) on unseen test tasks for Sparse Masked Attention policies at different sparsity levels $\alpha$ in each Procgen environment.}
    \label{fig:all-sparsitiy-levels}
\end{figure}

\end{document}